\documentclass{article}

\usepackage[preprint,nonatbib]{neurips_2023}

\usepackage[utf8]{inputenc} 
\usepackage[T1]{fontenc}    

\usepackage{url}            
\usepackage{booktabs}       
\usepackage{amsfonts}       
\usepackage{nicefrac}       
\usepackage{microtype}      
\usepackage{xcolor}         

\usepackage{times}
\usepackage{epsfig}
\usepackage{graphicx}
\usepackage{amsmath}
\usepackage{amssymb}
\usepackage{subcaption}
\usepackage{booktabs}
\usepackage{bm}
\usepackage{multirow}
\usepackage{bbding}
\usepackage[linesnumbered,ruled,vlined]{algorithm2e}

\usepackage{caption}
\usepackage{hyperref}       

\title{Distribution-Flexible Subset Quantization for Post-Quantizing Super-Resolution Networks}

\author{Yunshan Zhong$^{1,2}$, Mingbao Lin$^3$, Jingjing 
 Xie$^2$, Yuxin Zhang$^2$, Fei Chao$^{1,2}$, Rongrong Ji$^{1,2,4}$\thanks{Corresponding Author: rrji@xmu.edu.cn}\\
$^1$Institute of Artificial Intelligence, Xiamen University\\
$^2$MAC Lab, School of Informatics, Xiamen University \, \\ $^3$Tencent Youtu Lab \quad $^4$Peng Cheng Laboratory \\
{\tt\small zhongyunshan@stu.xmu.edu.cn, linmb001@outlook.com, 19119558205@163.com,} \\ 
{\tt\small yuxinzhang@stu.xmu.edu.cn, fchao@xmu.edu.cn,  rrji@xmu.edu.cn}
}

\begin{document}

\maketitle

\begin{abstract}
This paper introduces Distribution-Flexible Subset Quantization (DFSQ), a post-training quantization method for super-resolution networks. Our motivation for developing DFSQ is based on the distinctive activation distributions of current super-resolution models, which exhibit significant variance across samples and channels. To address this issue, DFSQ conducts channel-wise normalization of the activations and applies distribution-flexible subset quantization (SQ), wherein the quantization points are selected from a universal set consisting of multi-word additive log-scale values.
To expedite the selection of quantization points in SQ, we propose a fast quantization points selection strategy that uses $K$-means clustering to select the quantization points closest to the centroids. Compared to the common iterative exhaustive search algorithm, our strategy avoids the enumeration of all possible combinations in the universal set, reducing the time complexity from exponential to linear.
Consequently, the constraint of time costs on the size of the universal set is greatly relaxed.
Extensive evaluations of various super-resolution models show that DFSQ effectively retains performance even without fine-tuning. For example, when quantizing EDSR$\times$2 on the Urban benchmark, DFSQ achieves comparable performance to full-precision counterparts on 6- and 8-bit quantization, and incurs only a 0.1 dB PSNR drop on 4-bit quantization. Code is at \url{https://github.com/zysxmu/DFSQ}

\end{abstract}

\section{Introduction}

Image super-resolution (SR) is a fundamental low-level computer vision task that aims to restore high-resolution (HR) images from low-resolution input images (LR). Due to the remarkable success of deep neural networks (DNNs), DNNs-based SR models have become a \textit{de facto} standard for SR task~\cite{lim2017enhanced,zhang2018residual,dong2014learning,kim2016accurate,zhang2018image}.
However, the astonished performance of recent SR models typically relies on increasing network size and computational cost, thereby
limiting their applications, especially in resource-hungry devices such as smartphones.
Therefore, compressing SR models has gained
extensive attention from both academia and industries. Various network compressing techniques have been investigated to realize model deployment~\cite{lin2020hrank,krishnamoorthi2018quantizing,hinton2015distilling,han2015learning}.

Among these techniques, network quantization, which maps the full-precision weights and activations within networks to a low-bit format, harvests favorable interest from the SR community for its ability to reduce storage size and computation cost simultaneously~\cite{wang2021fully,hong2022daq,jiang2021training,xin2020binarized,ma2019efficient,li2020pams,hong2022cadyq,zhong2022dynamic}.
For example, \cite{xin2020binarized,ma2019efficient} quantize the SR models using binary quantization, and \cite{li2020pams,hong2022cadyq,zhong2022dynamic} quantize SR models to low-bit such as 2, 3, and 4-bit. 
Despite notable progress, current methods have to retrain quantized SR models on the premise of access to the entire training set, known as quantization-aware training (QAT). 
In real-world scenarios, however, acquiring original training data is sometimes prohibitive due to privacy, data transmission, and security issues. 
Besides, the heavy training and energy cost also prohibits its practical deployment.
Post-training quantization (PTQ) methods, which perform quantization with only a small portion of the original training set, require no or a little retraining, by nature can be a potential way to solve the above problems~\cite{Jeon2022MrBiQ,fang2020post,li2021brecq,Upordown,BitSplitStitching}. However, current PTQ methods mostly are designed for high-level vision tasks, a direct extension of which to SR models is infeasible since low-level models comprise different structures~\cite{li2020pams,zhong2022dynamic}.

\begin{figure*}[!t]
\centering
\subfloat[
{\centering \label{fig:insight-EDSRx4_channel}}]
{
\centering\includegraphics[width=0.8\linewidth]{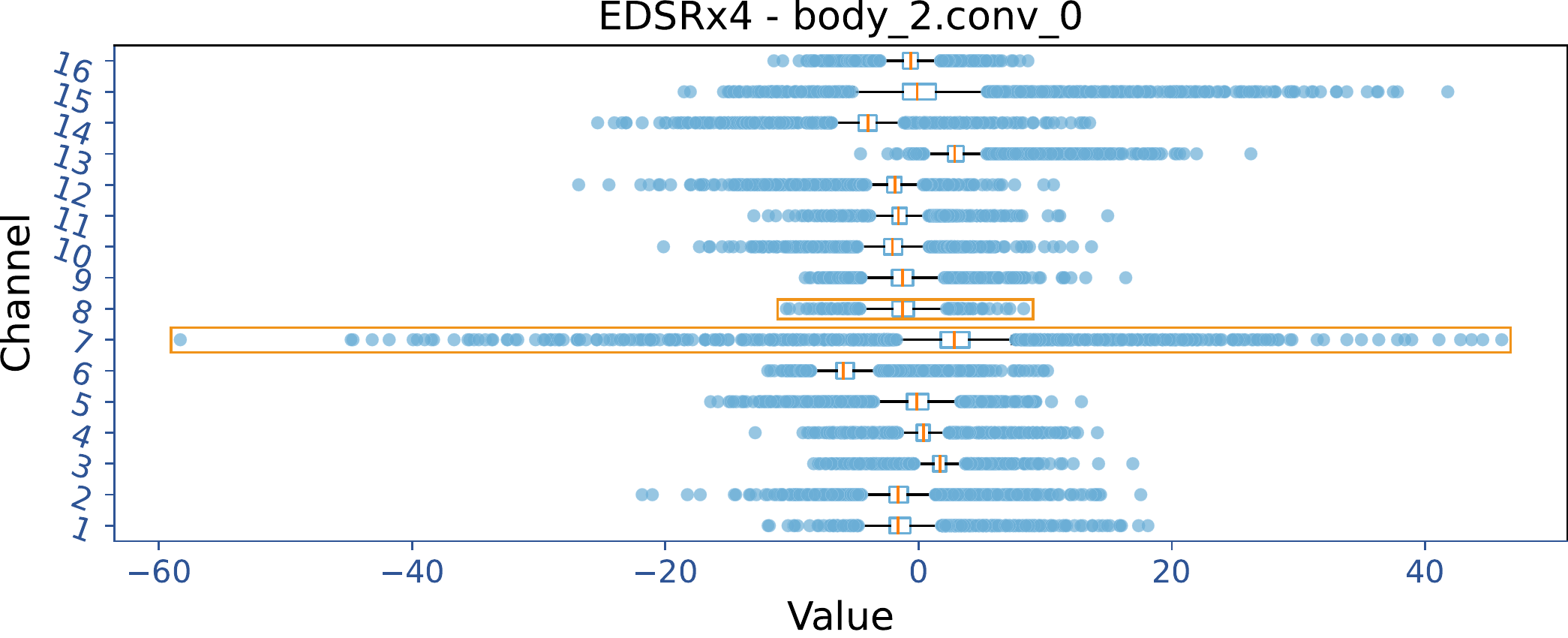} \hfill
}
\\
\subfloat[
{\centering \label{fig:insight-EDSRx4_sample}}]
{
\centering\includegraphics[width=0.8\linewidth]{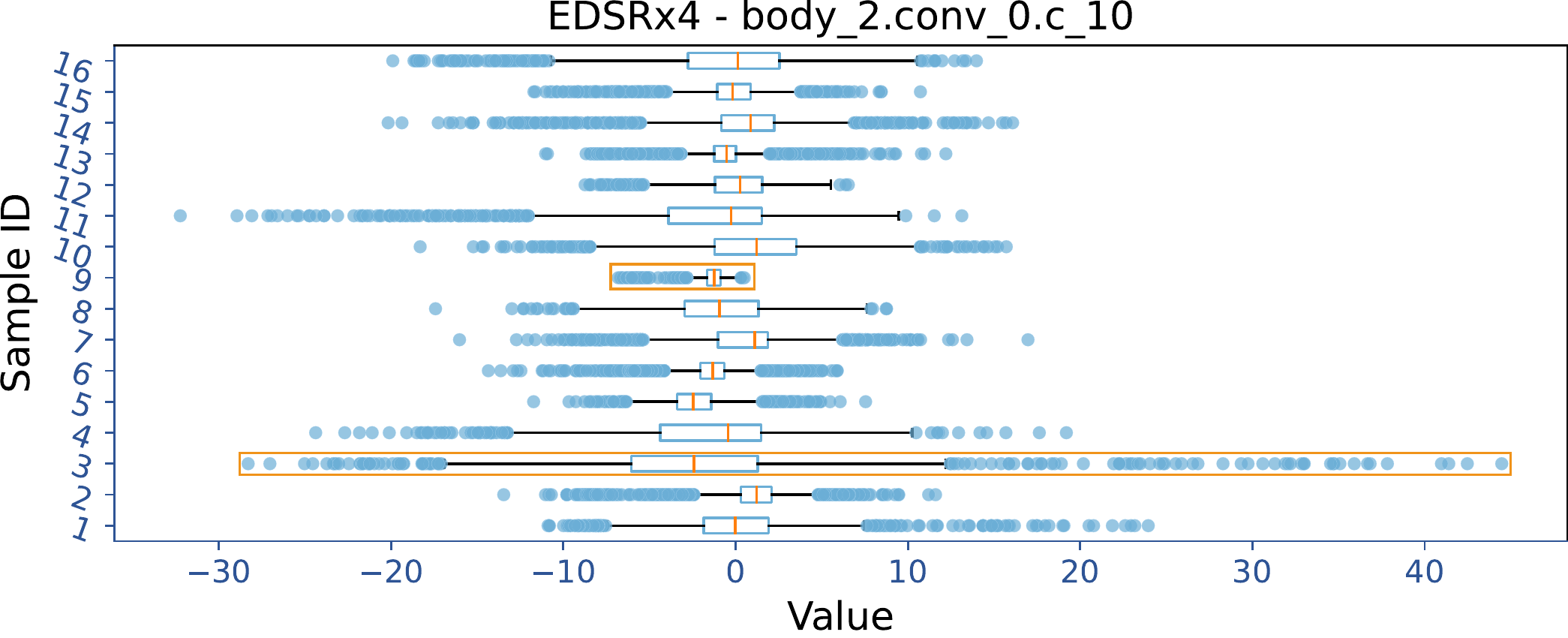} \hfill
}
\caption{Variation of activation distributions of EDSR$\times$4~\cite{lim2017enhanced}. We show activation distributions of different channels given the same input (a), and of different inputs given the same channel (b).  The orange box depicts the data with the maximum discrepancy.
}
\label{fig:insight}
\end{figure*}

More precisely, SR models usually remove all or most batch normalization (BN) layers since they reduce scale information within activations, which, as a wide consensus, is crucial to the performance of SR models~\cite{lim2017enhanced,zhang2018residual,zhang2018image,haris2018deep}.
Unfortunately, the main obstacle of quantization also comes from the removal of BN since it leads to high activation variations such as considerable distribution discrepancy among different channels of the same sample and among the same channel of different samples.
Taking Fig.\,\ref{fig:insight-EDSRx4_channel} as an example, for given a sample, the activation distributions among different channels vary a lot. At first, it can be seen the interquartile range (IQR) of channel\_7 and channel\_8 differs a lot. The IQR of the former is greater than 0 while the latter is less than 0. Second, the outlier distribution of channel\_7 is far wide than that of channel\_8, causing the activation range to differ by 5$\times$$\sim$6$\times$.
In addition, as illustrated in Fig.\,\ref{fig:insight-EDSRx4_sample}, different samples present significant distribution discrepancies in channel\_10. 
In particular, sample\_3 manifests 10$\times$ more IQR than that of sample\_9. Also, the activation of sample\_3 ranges from -30 $\sim$ 48, while sample\_9 only ranges from -8 $\sim$ 0. 
The high variance of activation across channels and samples makes it difficult to solve with current PTQ methods, which is also experimentally demonstrated in Sec.\,\ref{sec:quantitative}.

In this paper, we propose a Distribution-Flexible Subset Quantization (DFSQ) method to handle such highly variational activations. 
Specifically, considering the high variance among samples and channels, we first perform channel-wise normalization and then conduct distribution-flexible and hardware-friendly subset quantization (SQ)~\cite{Oh2022Nonuniform} to quantize the normalized activations.
The normalization comprises two consecutive on-the-fly operations including subtracting the mean and dividing by the maximum absolute value for each channel of each sample. As a result, the range is normalized to -1 $\sim$ 1 whatever the input sample and the activation channel.
Given the non-uniformity of normalized distribution, we suggest adopting the very recent subset quantization (SQ)~\cite{Oh2022Nonuniform}, which aims to find the best quantization points from a universal set that consists of log-scale values. 
However, the search of quantization points in \cite{Oh2022Nonuniform} involves an iterative exhaustive search algorithm that makes the time costs exponential \emph{w.r.t.} the size of the universal set, resulting in prohibitive time overhead and limiting the size of the universal set.
Therefore, we introduce a fast quantization points selection strategy to speed up the selection of quantization points in SQ.
In particular, we perform bit-width related $K$-means clustering at first. Then, from a given universal set, we select the quantization points closest to $K$ centroids. 
Our strategy circumvents the enumeration of all possible combinations of quantization points in the universal set, and thereby reduces the time complexity from exponential to linear. Consequently, the limitation of time costs on the size of the universal set is greatly relaxed.

Extensive evaluations of two well-known SR models including EDSR~\cite{lim2017enhanced} and RDN~\cite{zhang2018residual} on four benchmark datasets demonstrate the effectiveness of the proposed DFSQ. Notably, without any fine-tuning, DFSQ obtains comparable performance to the full-precision counterparts in high-bit cases such as 8 and 6-bit. For low-bit cases such as 4-bit, DFSQ is still able to greatly retain the performance. For example, on Urban100, DFSQ obtains 31.609 dB PSNR for EDSR$\times$2, only incurring less than 0.1 dB drops.

\section{Related Work}

\subsection{Single Image Super Resolution}
Along with the huge success of deep neural networks on many computer vision tasks, DNN-based SR models also obtain great performance increases and have dominated the field of image super-resolution.
As a pioneer, Chao \emph{et al}.~\cite{dong2014learning} for the first time proposed an end-to-end SRCNN to learn the mapping relationship between LR and HR images.
VDSR~\cite{kim2016accurate} further improves performance by increasing network depth.
Afterward, skip-connection based blocks~\cite{ledig2017photo,tong2017image} are extensively adopted by the subsequent studies~\cite{lim2017enhanced,zhang2018residual} to alleviate the gradient vanishing issue and retain image details.
For better performance, researchers introduce many complex structures to construct SR models such as channel attention mechanism~\cite{zhang2018image,magid2021dynamic}, non-local attention~\cite{mei2020non-local,mei2021non-local}, and transformer-based block~\cite{liang2021swinir,zamir2022restormer}.
With the increasing demand for the deployment of SR models on resource-limited devices, many studies aim to design lightweight network architectures.
DRCN~\cite{kim2016deeply} and DRRN~\cite{tai2017image} both adopt the recursive structure to increase the depth of models while reducing the model size. 
Some studies design modules to substitute for the expensive up-sampling operation. FSRCNN~\cite{dong2016accelerating} introduces a de-convolutional layer, and ESPCN~\cite{shi2016real} instead devises a sub-pixel convolution module.
Many other studies utilize the efficient intermediate feature representation~\cite{lai2017deep,ahn2018fast,hui2019lightweight,luo2020latticenet} or network architecture search~\cite{oh2022attentive}.
%

\subsection{Quantized SR Models}

Network quantization enjoys the merit of both reducing storage size and efficient low-bit operations and thereby harvesting ever-growing attention~\cite{wang2021fully,hong2022daq,jiang2021training,xin2020binarized,ma2019efficient,li2020pams,hong2022cadyq,zhong2022dynamic}.
Ma \emph{et al.}~\cite{ma2019efficient}  proposed binary quantization for the weights within SR models. Following them, BAM~\cite{xin2020binarized} and BTM~\cite{jiang2021training} further binarize the activation of SR models. They introduce multiple feature map aggregations and skip connections to reduce the sharp performance drops caused by the binary activation.
Other than binary quantization, many studies focus on performing low-bit quantization~\cite{wang2021fully,hong2022daq,li2020pams,hong2022cadyq,zhong2022dynamic}.
Li \emph{et al.}~\cite{li2020pams} found unstable activation ranges and proposed a symmetric layer-wise linear quantizer, where a learnable clipping value is adopted to regulate the abnormal activation.
Moreover, a knowledge distillation loss is devised to transfer structured knowledge of the full-precision model to the quantized model.
Wang \emph{et al.}~\cite{wang2021fully} designed a fully-quantization method for SR models, in which the weight and activation within all layers are quantized with a symmetric layer-wise quantizer equipped with a learnable clipping value.
Zhong \emph{et al.}~\cite{zhong2022dynamic} observed that the activation exhibit highly asymmetric distributions and the range magnitude drastically varies with different input images. They introduced two learnable clipping values and a dynamic gate to adaptively adjust the clipping values.
In~\cite{hong2022cadyq}, a dynamic bit-width adjustment network is introduced for different input patches that have various structure information.
%

\section{Method}

QAT usually trains the quantized network for many epochs, by accessing the entire training set, to gradually accommodate the quantization effect~\cite{LSQ}.
Differently, PTQ is confined to a small portion of the original training set, leading to a severe over-fitting issue~\cite{li2021brecq}. Thus, the key to PTQ has drifted to fitting the data distribution. Below, we first demonstrate the obstacle in performing PTQ for SR models lies in the high variance activation distributions. Then, we introduce the subset quantization and a corresponding fast selection strategy.

\subsection{Observations}

It is a wide consensus that the removal batch normalization layer in SR models improves the quality of output HR images~\cite{lim2017enhanced,zhang2018residual,zhang2018image,haris2018deep}. 
Unfortunately, as discussed in many previous studies~\cite{hong2022daq,zhong2022dynamic,li2020pams}, the removal of the BN layer creates the obstacle for quantization since the resulting activations of high variance make low-bit networks hard to fit. 
In particular, the high-variance activations are two folds: 1) considerable distribution discrepancy in different channels for a given input sample; 2) considerable distribution discrepancy in different samples for a given channel.

Fig.\,\ref{fig:insight} presents the example of activation distributions within EDSR$\times$4~\cite{lim2017enhanced}.
Specifically, Fig.\,\ref{fig:insight-EDSRx4_channel} presents the activation distributions of different channels given the same sample. It can be seen that the distribution exhibits significant discrepancy. For example, the interquartile range of channel\_7 is greater than 0 while channel\_8 is less than 0.  
Moreover, the range of channel\_7 is far wide than that of channel\_8, resulting in a range difference by 5$\times$$\sim$6$\times$.
Fig.\,\ref{fig:insight-EDSRx4_sample} presents the activation distribution of different input samples given the same channel. As it shows, given the same channel\_10, extreme discrepancies between the distribution of different samples are revealed.
Taking sample\_3 and sample\_9 as the example, the IQR of the former ranges from -6$\sim$1.5, while the latter ranges from -1.64$\sim$-0.8, almost 10$\times$ difference. Also, their range differs a lot. The activation of sample\_3 ranges from -30$\sim$48, while sample\_9 only ranges from -8$\sim$0, still resulting in a 10$\times$ difference.
Therefore, the high-variance activation distribution is reflected by extreme discrepancy among different channels and different samples.

To handle such activation distributions, previous SR quantization methods rely on QAT to gradually adjust network weights to accommodate the quantization effect~\cite{LSQ}. However, PTQ hardly succeeds in this manner since the availability of partial data easily causes over-fitting issue~\cite{li2021brecq}. Moreover, the highly nonuniform distribution also makes the common linear uniform quantization adopted by current PTQ methods hard to fit the original distribution~\cite{Li2020Additive}, which is also observed from our experimental results in Sec.\,\ref{sec:quantitative}.
Therefore, the core is to find a suitable quantizer that can well fit the distribution as much as possible.

\subsection{Distribution-Flexible Subset Quantization for Activation}

\subsubsection{Quantization Process}
We denote a full-precision feature map as $X \in R^{B \times C \times H \times W}$, where $B, C, H, W$ respectively denote mini-batch size, channel number, height, and width of the feature maps. Considering that the activation distributions vary quite a lot across samples and channels, we choose to perform quantization for each channel of each sample, denoted as $X_{i, c} \in R^{H \times W}$, where $X_{i, c}$ denotes the $c$-th feature map (channel) of the $i$-th input sample. We suggest conducting normalization on-the-fly at first to scale the activation of $X_{i, c}$ as:
\begin{equation}
X^n_{i, c} = f_n(X_{i, c}) = \frac{X_{i, c} - \mu_{i, c}}{\mathcal{M}_{i, c} - \mu_{i, c}}, 
\label{norm}
\end{equation}
where $\mu_{i, c}$ and $\mathcal{M}_{i, c}$ denote the mean value and maximum absolute value of $X_{i, c}$. The superscript ``$n$'' represents normalization.

After normalization, the activation is scaled to -1$\sim$1 whatever the input and channel are, thereby facilitating the following quantization:
\begin{equation}
X^q_{i, c} = Q(X^n_{i, c}),
\label{quant}
\end{equation}
$Q(\cdot)$ denotes the quantizer, which is elaborated in the next subsection. The superscript ``$q$'' denotes quantized results.
Then, the de-quantized activation can be obtained by conducting de-normalization after getting $X^q_{i, c}$:
\begin{equation}
\bar{X}_{i, c} = f_n^{-1}(X^q_{i, c}) = X^q_{i, c} \cdot \mathcal{M}_{i, c} + \mu_{i, c}. 
\label{de-norm}
\end{equation}

Note that, the overhead of de-normalization can be reduced by quantizing $\mu_{i, c}$ and $\mathcal{M}_{i, c}$ to a low-bit format, which has already been studied in~\cite{dai2021vs,hong2022daq}.

\begin{figure*}[!t]
\centering
\includegraphics[width=0.95\linewidth]{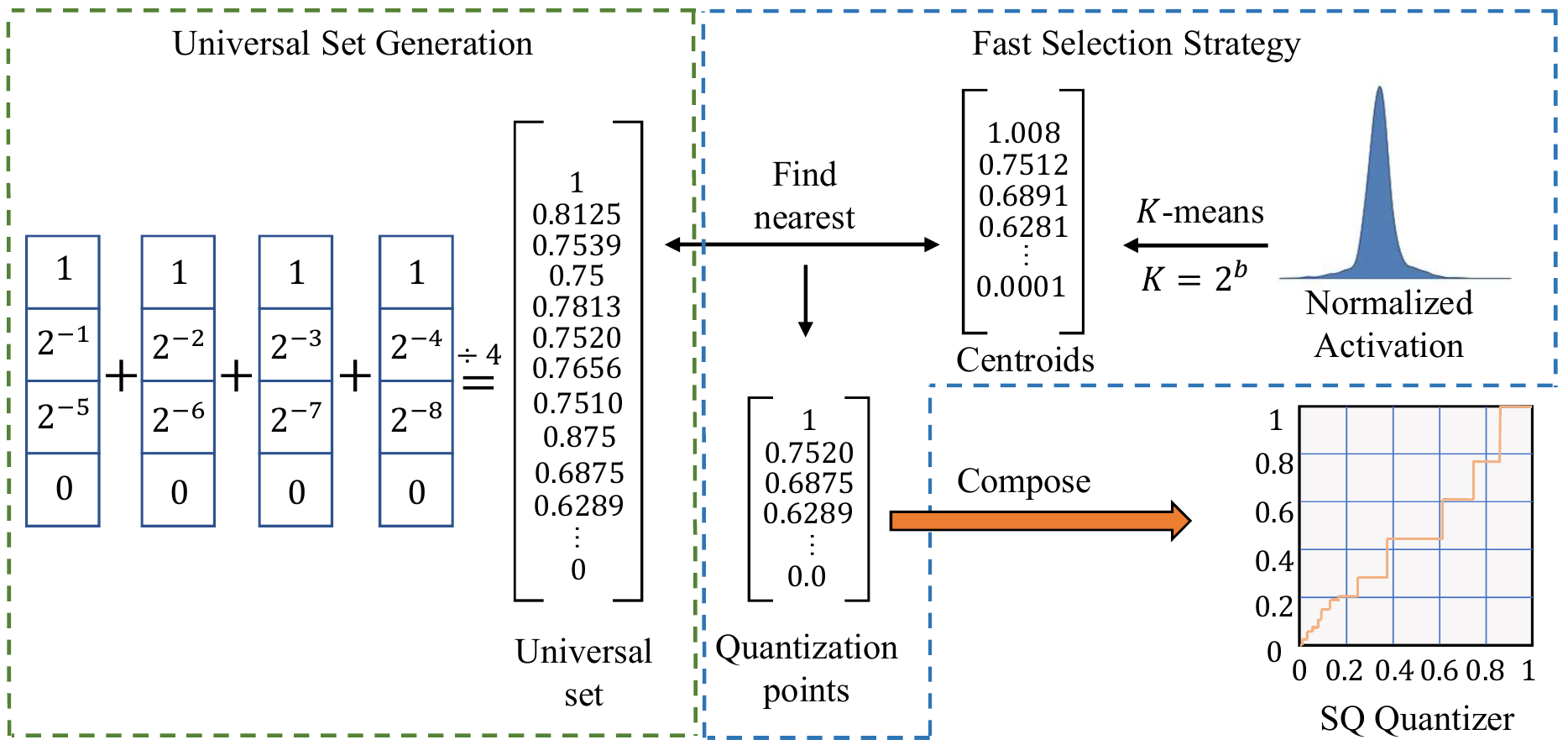} 
\caption{The framework of the proposed distribution-flexible subset quantization.}
\label{fig:framework}
\end{figure*}

\subsubsection{Subset Quantization}

In this subsection, we elaborate on the aforementioned quantizer $Q(\cdot)$. Despite that the normalized activations conform to the same range across different channels and samples, they are still featured with high non-uniformity. Such erratic distributions are hard to be fitted by the common linear uniform quantization~\cite{Li2020Additive}.

Therefore, we suggest the very recent distribution-flexible and hardware-friendly subset quantization (SQ)~\cite{Oh2022Nonuniform}.
Specifically, SQ aims to find the best quantization points from a predefined universal set that usually consists of the additive of multi-word log-scale values~\cite{Li2020Additive,oh2021automated,Lee2019succe}.
Given a universal set $\Phi_u$, bit-width $b$, and an input value $x$, the quantizer $Q(\cdot)$ is defined as:
\begin{equation}
\begin{split}
Q(x) & = \mathop{\arg\min}\limits_{p \in \Phi_s} | x - p|, \\
s.t. \quad \Phi_s & = \{p_i \in \Phi_u | i = 1, \cdots, 2^b\},
\label{act-quantizer}
\end{split}
\end{equation}
where $\Phi_s$ is the set of selected quantization points.
Therefore, the key steps of SQ are the universal set generation and the quantization points selection. 

\textbf{Universal Set Generation}.
The universal set $\Phi_u$ should contain adequate candidate values to represent any given input distribution~\cite{Oh2022Nonuniform}. 
The universal set used in our paper is presented in the left part of Fig.\,\ref{fig:framework}. Specifically, given four word sets, each of which contains four elements that are either zero or log-scale values. The universal set is obtained by averaging all possible combinations of the elements from each word set. For example, the value 0.8125 in the universal set is obtained by averaging the sum of 0 from the first word set, $2^{-2}$ from the second word set, and 1s from the third and fourth word sets. As a result, we can obtain a universal set that consists of many non-negative values.
For distribution with negative values, it needs negative values.
To do this, we add the negative values by changing the sign of non-negative values. For instance, if the value 0.8125 is present, its corresponding negative value, -0.8125, will also be added to the universal set.
After duplicate removal, a total of 107 values are given as the universal set.
Moreover, since each value of the universal set consists of four values that are either zero or $2^k$ and a division of 4, we can fuse the division of 4 into the elements of each word set. As a result, the multiplication between quantized activation and quantized weight only requires four shifters and one adder, which is hardware-friendly.

\textbf{Quantization Points Selection}.
The common selection strategy of quantization points involves an iterative exhaustive search algorithm, where all possible combinations in the universal set are exhaustively checked to find out the one that minimizes the quantization loss. However, such a strategy incurs intolerable time costs as the size of the universal set increases. In particular, the number of all possible combinations is $C_{n}^{2^b} = \frac{n!}{2^b!(n-2^b)!} $, where $b$ is the bit-width and $n$ is the size of the universal set. It can be seen the increase of combinations is an exponential growth \emph{w.r.t.} the size of the universal set. For example, given the 4-bit case and the universal set defined above, the total number of combinations is $C_{107}^{2^4} = 4.336 \times 10^{18} $. 
Therefore, the size of the universal set is limited and a fast selection strategy is necessary.

\subsubsection{Fast Selection Strategy}

We then introduce a fast quantization points selection strategy to speed up the selection of quantization points in SQ.
We are mainly inspired by the $K$-means algorithm which can be viewed as a solution for the quantization loss minimization problem with a given distribution and bit-width setting~\cite{wiki-quantization}.
In particular, given $X^n_{i, c}$, we perform $K$-means clustering at first by setting $K=2^b$. To avoid the local optimum, in practice, we perform $K$-means by 3 times and select the results with the minimum sum of squared errors (SSE):
\begin{equation}
\begin{split}
\Phi_{\mu} & = \mathop{\arg\min}\limits_{\Phi^i_{\mu} \in \{\Phi^1_{\mu}, \Phi^2_{\mu}, \Phi^3_{\mu}\}} \text{SSE}_{\Phi^i_{\mu}},
\label{k-means}
\end{split}
\end{equation}
where $\text{SSE}_{\Phi^i_{\mu}} = \sum_{x \in C^i_{j}} \| x - \mu^i_{j} \|^2$,  $\Phi^i_{\mu}$ denotes the centroids set of $i$-th trial of $K$-means, $C^i_{j}$ denotes the $j$-th cluster of the $i$-th trial, and $\mu^i_{j}$ denotes the centroid of $C^i_{j}$.
Then, from a given universal set, the quantization points set $\Phi_s$ is built by selecting these points closest to $K$ centroids of $\Phi_{\mu}$. 

The time complexity of $K$-means is $\mathcal{O}(NTK)$, where $N = H \times W$ is the total number of elements in $X^n_{i, c}$, $T$ is the number of iterations in the clustering process, and $K=2^b$~\cite{hartigan1979algorithm}. Also, the selection of quantization points closest to $K$ centroids only requires $\mathcal{O}(|\Phi_u|2^b)$. Note that the maximum bit-width $b$ generally is 8, which gives the $K=256$ at most. Thus, it is safe to say the time complexity of our fast selection strategy is linear.
Note that, in practice, we can utilize the multiprocessing mechanism to process the feature map of each channel in parallel, enabling more efficient use of computing resources.

In summary, the time complexity of our strategy linearly depends on the $N, K, T$ and the size of the universal set. Compared with the previous iterative exhaustive search algorithm, the time complexity is reduced from exponential to linear. 
The cumbersome drudgery of enumerating all possible combinations is avoided and therefore the limitation of time costs on the size of the universal set is greatly relaxed.

\subsection{Weight Quantization}

For weight quantization, we adopt kernel-wise linear uniform quantization. Given the weight $W \in R^{C_{out} \times C_{in} \times K \times K}$, where $C_{out}, C_{in}, K$ denote output channel number, input channel number, and kernel size, respectively. For a kernel $W_k$, the quantizer is defined as:
\begin{equation}
W^q_k = round(\frac{W}{s}) + Z, \quad s = \frac{u_w - l_w}{2^{b}-1}, \quad Z=round(\frac{-l_w}{s}),
\label{weight-quant}
\end{equation}
where $b, l_w, u_w, s, Z$ denote bit-width, weight minimum, weight maximum, step size, and zero-point integer corresponding to the full-precision $0$ respectively. The de-quantized value is obtained by:
\begin{equation}
\bar{W}_k = s \cdot (W^q_k - Z).
\label{weight-dequant}
\end{equation}

\section{Experimentation}

\subsection{Implementation Details}

The quantized SR models include two classical EDSR~\cite{lim2017enhanced} and RDN~\cite{zhang2018residual}. 
For each SR model, we evaluate two upscaling factors of $\times$2 and $\times$4 and perform 8-, 6-, 4-, and 3-bit quantization, respectively. The calibration dataset contains 32 images random sampled from the training set of DIV2K~\cite{timofte2017ntire}. 
The models are tested on four standard benchmarks including Set5~\cite{bevilacqua2012low}, Set14~\cite{ledig2017photo}, BSD100~\cite{martin2001database} and Urban100~\cite{huang2015single}. 
For the compared method, we adopt the Min-Max linear uniform quantization and two recent optimization-based methods including BRECQ~\cite{li2021brecq} and QDROP~\cite{wei2022qdrop}.
We report the PSNR and SSIM~\cite{wang2004image} over the Y channel as the metrics. 

The full-precision models and compared methods are implemented based on the official open-source code.
Following \cite{li2020pams,zhong2022dynamic}, we quantize both weights and activations of the high-level feature extraction module of the quantized models. The low-level feature extraction and reconstruction modules retain the full-precision.
All experiments are implemented with PyTorch~\cite{paszke2019pytorch}

\begin{table*}[!t]
\centering
\caption{Effect of different components in our paper. ``Cha.'': channel-wise activation quantization. ``Norm.'': normalization. The results are obtained by quantizing EDSR$\times$4 to 4-bit and the PSNR/SSIM is reported as the metrics.}
\begin{tabular}{cc|cccc} \toprule[1.25pt]
\multicolumn{2}{c|}{Components} & \multicolumn{4}{c}{Results}       \\ \hline \hline
Cha.      & Norm.      & Set5~\cite{bevilacqua2012low}  & Set14~\cite{ledig2017photo} & BSD100~\cite{martin2001database} & Urban100~\cite{huang2015single} \\ \hline
&  & - & - & - & -  \\
\checkmark    &    &  - & - & - & - \\
  &  \checkmark    & 31.543/0.8790 & 28.262/0.7675  & 27.358/0.7247   & 25.601/0.7586 \\
\checkmark &  \checkmark &  31.689/0.8835 & 28.339/0.7728  & 27.409/0.7295    &  25.723/0.7707\\\bottomrule[0.75pt]       
\end{tabular}
\label{tab:ablation}
\end{table*}

\begin{table*}[!t]
\centering
\setlength\tabcolsep{4.5pt}
\caption{PSNR/SSIM results of the compared baseline and our DFSQ in quantizing EDSR~\cite{lim2017enhanced} of scale $\times$2 and $\times$4. Results of the full-precision model are presented below the dataset name.}
\begin{tabular}{ccccccc}
\toprule[1.25pt]
Model                                                               & Dataset                   &Bit & BRECQ~\cite{li2021brecq}        & QDROP~\cite{wei2022qdrop} & Min-Max                     & \textbf{DFSQ(Ours)}          \\ \hline \hline
\multirow{16}{*}{\begin{tabular}[c]{@{}c@{}}EDSR\\ $\times$2\end{tabular}} & \multirow{5}{*}{\begin{tabular}[c]{@{}c@{}}Set5~\cite{bevilacqua2012low}\\ 37.931/0.9604 \end{tabular}}     & 8         & 37.921/0.9603 & 37.926/0.9603  & 37.926/0.9603             & \textbf{37.928/0.9603} \\
&                           & 6         &   37.865/0.9597      &  37.882/0.9598      & 37.905/0.9601           & \textbf{37.927/0.9603}        \\
&                           & 4         &  37.499/0.9564      &   37.370/0.9572     & 37.497/0.9559           & \textbf{37.832/0.9599}        \\
&                           & 3         &    36.845/0.9500      &   36.603/0.9526   &  36.199/0.9383     & \textbf{37.382/0.9567}        \\ \cline{2-7}  
& \multirow{4}{*}{\begin{tabular}[c]{@{}c@{}}Set14~\cite{ledig2017photo}\\ 33.459/0.9164 \end{tabular}}    & 8         &   33.457/0.9163      &    33.457/0.9163   & 33.451/0.9164           & \textbf{33.459/0.9164}        \\
&                           & 6         &  33.419/0.9157       &  33.398/0.9159       & 33.436/0.9161           & \textbf{33.455/0.9164}        \\
&                           & 4         &   33.138/0.9123    &   32.947/0.9123      & 33.229/0.9122           & \textbf{33.399/0.9159}        \\
&                           & 3         &    32.714/0.9054     &  32.459/0.9074     &       32.477/0.8955      & \textbf{33.068/0.9113}        \\ \cline{2-7}  
& \multirow{4}{*}{\begin{tabular}[c]{@{}c@{}}BSD100~\cite{martin2001database}\\ 32.102/0.8987 \end{tabular}}   & 8         &  32.098/0.8986      &    32.099/0.8986     & 32.100/0.8987           & \textbf{32.101/0.8987}        \\
 &                           & 6         &    32.066/0.8979     &   32.060/0.8980      &    32.089/0.8984       & \textbf{32.099/0.8986}        \\
 &                           & 4         &     31.829/0.8939    &     31.710/0.8939   & 31.911/0.8941          & \textbf{32.060/0.8981}        \\
 &                           & 3         &   31.475/0.8866       & 31.324/0.8885     & 31.263/0.8764       & \textbf{31.824/0.8939}        \\ \cline{2-7}  
 & \multirow{4}{*}{\begin{tabular}[c]{@{}c@{}}Urban100~\cite{huang2015single}\\ 31.709/0.9248 \end{tabular}} & 8         &     31.698/0.9246    &    31.683/0.9245     & 31.663/0.9245            & \textbf{31.707/0.9247}        \\
 &                           & 6         &     31.588/0.9235    &   31.463/0.9228      & 31.642/0.9241          & \textbf{31.698/0.9246}        \\
 &                           & 4         &   30.874/0.9158      &    30.265/0.9111    & 31.367/0.9188          & \textbf{31.609/0.9236}        \\ 
 &                           & 3         &     30.106/0.9041     &  29.407/0.8989     &   30.395/0.8977          & \textbf{30.972/0.9137}        \\  \hline \hline
\multirow{16}{*}{\begin{tabular}[c]{@{}c@{}}EDSR\\ $\times$4\end{tabular}} & \multirow{4}{*}{\begin{tabular}[c]{@{}c@{}}Set5~\cite{bevilacqua2012low}\\ 32.095/0.8938 \end{tabular}}     & 8         & 32.088/0.8935 & 32.089/0.8936 & 32.087/0.8936 &      \textbf{32.090/0.8937} \\
&                           & 6         &      32.018/0.8909  &  31.996/0.8911   & 32.056/0.8925           & \textbf{32.079/0.8933}        \\
&                           & 4         &   31.287/0.8722      &   31.103/0.8715      & 31.364/0.8687           & \textbf{31.755/0.8855}        \\
&                           & 3         &   30.164/0.8342      &   30.286/0.8478   &     29.150/0.7580        & \textbf{30.757/0.8489}        \\ \cline{2-7}  
& \multirow{4}{*}{\begin{tabular}[c]{@{}c@{}}Set14~\cite{ledig2017photo}\\ 28.576/0.7813 \end{tabular}}    & 8         &   28.566/0.7809      &   28.566/0.7810      &  28.566/0.7809          & \textbf{28.568/0.7810}        \\
&                           & 6         &  28.516/0.7788       &  28.501/0.7788    & 28.549/0.7801           & \textbf{28.560/0.7807}        \\
&                           & 4         &  28.080/0.7635      &    27.922/0.7634     & 28.159/0.7617           & \textbf{28.397/0.7749}        \\
&                           & 3         &  27.396/0.7330       &  27.392/0.7446    &     26.723/0.6681        & \textbf{27.732/0.7431}        \\ \cline{2-7} 
& \multirow{4}{*}{\begin{tabular}[c]{@{}c@{}}BSD100~\cite{martin2001database}\\ 27.562/0.7355 \end{tabular}}   & 8         &   27.557/0.7352      &     27.557/0.7352     & 27.555/0.7351           & \textbf{27.558/0.7354}        \\
&                           & 6         &   27.507/0.7326      &  27.509/0.7330 & 27.547/0.7344           & \textbf{27.555/0.7351}        \\
&                           & 4         &  27.198/0.7184     &     27.153/0.7197     & 27.255/0.7168           & \textbf{27.430/0.7307}        \\
&                           & 3         &    26.717/0.6903     &   26.811/0.7032   &    26.117/0.6253         & \textbf{27.044/0.7074}        \\ \cline{2-7}  
& \multirow{4}{*}{\begin{tabular}[c]{@{}c@{}}Urban100~\cite{huang2015single}\\ 26.035/0.7848 \end{tabular}} &  8         &  26.018/0.7843      &   26.002/0.7841    & 26.014/0.7844             & \textbf{26.025/0.7845}        \\
&                           & 6         &   25.907/0.7801      &  25.849/0.7791    & 25.997/0.7831          &  \textbf{26.020/0.7840}      \\
&                           & 4         &   25.291/0.7543      &   25.044/0.7485       &   25.588/0.7595    & \textbf{25.769/0.7736}        \\
&                           & 3         &   24.560/0.7124      &  24.460/0.7188    &      24.287/0.6520       & \textbf{24.987/0.7249}        \\ \bottomrule[0.75pt]
\end{tabular}
\label{EDSR}
\end{table*}

\begin{table*}[ht]
\centering
\setlength\tabcolsep{4.5pt}
\caption{PSNR/SSIM results of the compared baseline and our DFSQ in quantizing RDN~\cite{lim2017enhanced} of scale $\times$2 and $\times$4.}
\begin{tabular}{ccccccc}
\toprule[1.25pt]
Model                                                               & Dataset                   &Bit & BRECQ~\cite{li2021brecq}        & QDROP~\cite{wei2022qdrop} & Min-Max                     & \textbf{DFSQ(Ours)}          \\ \hline \hline
\multirow{16}{*}{\begin{tabular}[c]{@{}c@{}}RDN\\ $\times$2\end{tabular}} & \multirow{4}{*}{\begin{tabular}[c]{@{}c@{}}Set5~\cite{bevilacqua2012low}\\ 38.053/0.9607\end{tabular}}     & 8         & 38.019/0.9603 & 38.020/0.9604 & 38.049/0.9606             & \textbf{38.053/0.9607} \\
&                           & 6         & 37.884/0.9588        & 37.873/0.9591        & 37.975/0.9599           & \textbf{38.042/0.9606}        \\
&                           & 4         & 37.143/0.9538        & 37.050/0.9540        & 37.172/0.9544           & \textbf{37.786/0.9593}        \\
&                           & 3         & 36.135/0.9469        & 36.000/0.9469        & 35.872/0.9464           & \textbf{37.125/0.9559} \\\cline{2-7}  
& \multirow{4}{*}{\begin{tabular}[c]{@{}c@{}}Set14~\cite{ledig2017photo}\\ 33.594/0.9174\end{tabular}}    & 8         & 33.576/0.9172        & 33.566/0.9172        & 33.560/0.9175           & \textbf{33.589/0.9174}        \\
&                           & 6         & 33.464/0.9158        &  33.417/0.9160        & 33.530/0.9168           & \textbf{33.585/0.9173}        \\
&                           & 4         & 32.949/0.9106        &  32.825/0.9101        & 33.136/0.9108           & \textbf{33.373/0.9154}        \\
&                           & 3         & 32.322/0.9032        & 32.158/0.9012        & 32.297/0.8996           & \textbf{32.896/0.9105} \\\cline{2-7}  
& \multirow{4}{*}{\begin{tabular}[c]{@{}c@{}}BSD100~\cite{martin2001database}\\ 32.197/0.8998\end{tabular}}   & 8         &  32.185/0.8995        & 32.187/0.8996        & 32.193/0.8998           & \textbf{32.195/0.8998}        \\
 &                           & 6         & 32.120/0.8982        &  32.115/0.8985        & 32.167/0.8991          & \textbf{32.184/0.8996}        \\
 &                           & 4         & 31.761/0.8932        & 31.689/0.8934        & 31.818/0.8913          & \textbf{32.043/0.8973}        \\
 &                           & 3         & 31.281/0.8859        & 31.191/0.8851        & 31.133/0.8782          & \textbf{31.675/0.8919} \\\cline{2-7}  
 & \multirow{4}{*}{\begin{tabular}[c]{@{}c@{}}Urban100~\cite{huang2015single}\\ 32.125/0.9286 \end{tabular}} & 8         & 32.088/0.9282        & 32.051/0.9282        &  32.014/0.9281            & \textbf{32.115/0.9285}        \\
 &                           & 6         & 31.827/0.9260        & 31.713/0.9257        & 31.975/0.9274          & \textbf{32.079/0.9281}        \\
 &                           & 4         & 30.681/0.9150        & 30.367/0.9128        &  31.418/0.9193          & \textbf{31.594/0.9217}        \\
 &                           & 3         &  29.532/0.8989        & 29.214/0.8930        & 30.086/0.8998           & \textbf{30.504/0.9080} \\ \hline \hline
\multirow{16}{*}{\begin{tabular}[c]{@{}c@{}}RDN\\ $\times$4\end{tabular}} & \multirow{4}{*}{\begin{tabular}[c]{@{}c@{}}Set5~\cite{bevilacqua2012low}\\ 32.244/0.8959\end{tabular}}     & 8         & 32.233/0.8953 & 32.230/0.8954 & 32.238/0.8956 &      \textbf{32.244/0.8959} \\
&                           & 6         & 32.148/0.8929        & 32.141/0.8930        & 32.191/0.8941           & \textbf{32.228/0.8955}        \\
&                           & 4         &  31.498/0.8801        & 31.341/0.8794        & 31.619/0.8798           & \textbf{31.932/0.8895}        \\
&                           & 3         & 30.509/0.8586        & 30.455/0.8602        & 30.430/0.8546           & \textbf{31.077/0.8718}        \\ \cline{2-7}  
& \multirow{4}{*}{\begin{tabular}[c]{@{}c@{}}Set14~\cite{ledig2017photo}\\ 28.669/0.7838\end{tabular}}    & 8         & 28.657/0.7834        & 28.650/0.7834        &  28.642/0.7835          & \textbf{28.663/0.7837}        \\
&                           & 6         & 28.582/0.7811        & 28.563/0.7812        &  28.616/0.7822           & \textbf{28.653/0.7834}        \\
&                           & 4         &  28.139/0.7701        & 28.032/0.7696        &  28.303/0.7701           & \textbf{28.471/0.7774}        \\
&                           & 3         & 27.516/0.7528        & 27.468/0.7537        & 27.621/0.7478           & \textbf{27.941/0.7616}        \\\cline{2-7} 
& \multirow{4}{*}{\begin{tabular}[c]{@{}c@{}}BSD100~\cite{martin2001database}\\  27.627/0.7379\end{tabular}}   & 8         &  27.620/0.7375        &  27.621/0.7376        & 27.618/0.7377           & \textbf{27.625/0.7378}        \\
&                           & 6         & 27.575/0.7354        & 27.572/0.7360        & 27.597/0.7365           & \textbf{27.616/0.7375}        \\
&                           & 4         & 27.305/0.7261        &  27.249/0.7268        & 27.367/0.7245           & \textbf{27.504/0.7326}        \\
&                           & 3         & 26.918/0.7123        & 26.916/0.7143        & 26.889/0.7037           & \textbf{27.158/0.7195}        \\\cline{2-7}  
& \multirow{4}{*}{\begin{tabular}[c]{@{}c@{}}Urban100~\cite{huang2015single}\\ 26.293/0.7924\end{tabular}} &  8         & 26.262/0.7916        & 26.245/0.7915        &  26.182/0.7904           & \textbf{26.292/0.7924}        \\
&                           & 6         &  26.116/0.7875        & 26.061/0.7870        & 26.157/0.7891           &  \textbf{26.266/0.7914}      \\
&                           & 4         & 25.448/0.7662        & 25.292/0.7636        & 25.789/0.7720           & \textbf{25.960/0.7780}        \\
&                           & 3         & 24.700/0.7351        & 24.590/0.7331        & 24.921/0.7340           & \textbf{25.156/0.7442}        \\\bottomrule[0.75pt]
\end{tabular}
\label{RDN}
\end{table*}

\subsection{Ablation Study}

The ablation study\footnote{The ablation study of different universal sets is presented in the supplementary materials.} of different components in our paper is presented in Tab.\,\ref{tab:ablation}. When utilizing both channel-wise activation and normalization, our DFSQ presents the best results. As shown in the results of the first row and the second row, the quantized model suffers from collapse if normalization is not applied, indicating the importance of normalization for the subset quantization.
The third row provides the results of not applying channel-wise quantization for activation, \emph{i.e.}, layer-wise activation quantization. It can be seen that channel-wise activation brings performance improvements. In particular, on Urban100, the quantized model presents 25.601 dB PSNR if not using channel-wise activation, while it is 25.732 dB PSNR if using channel-wise activation, demonstrating the effectiveness of handling each channel independently.

\subsection{Quantitative Results}
\label{sec:quantitative}

In this subsection, we provide quantitative results of EDSR and RDN across various bit-widths. The qualitative results are presented in the supplementary materials.

\subsubsection{EDSR}

Tab.\,\ref{EDSR} presents the quantitative results of EDSR$\times$2 and EDSR$\times$4. It can be seen that our DFSQ obtains the best performance across different datasets and bit-widths. 
Specifically, when performing high-bit PTQ, such as 8- and 6-bit quantization, our DFSQ achieves comparable performance to the full-precision counterpart.
For instance, on 8- and 6-bit EDSR$\times$2, DFSQ obtains 32.101 dB and 32.099 dB PSNR on BSD100, which only gives a drop of 0.001 dB and 0.003 dB compared with the full-precision model, respectively. Also, results of 8- and 6-bit EDSR$\times$4 on BSD100 demonstrate that DFSQ only incurs 0.004 dB and 0.007 dB PSNR drop, respectively.
It is worth emphasizing that the performance superiority of DFSQ exhibits as the bit-width goes down. Taking results of EDSR$\times$2 on Urban100 as the example, compared with the best result from other competitors, DFSQ obtains gains of 0.009 dB PSNR. For 6-, 4-, and 3-bit cases, DFSQ brings improvement of 0.056 dB, 0.242 dB, and 0.532 dB PSNR, respectively. Results of EDSR$\times$4 also provide a similar conclusion. For example, on Urban100, our DFSQ improves the PSNR by 0.007 dB, 0.023 dB, 0.181 dB, and 0.427 dB for For 8-, 6-, 4-, and 3-bit cases, respectively.
Moreover, despite fine-tuning the weights, optimization-based BRECQ and QDROP exhibit lower performance than the simple Min-Max at most bit-widths, indicating they suffer from the over-fitting issue. In contrast, our DFSQ does not need any fine-tuning, and still achieves stable superior performance across all bit-widths.

\subsection{RDN}

Quantitative results of RDN are presented in Tab.\,\ref{RDN}. As can be seen, DFSQ obtains the best performance over different bit-widths and datasets.
For the high-bit cases, DFSQ provides comparable performance to the full-precision model. For example, on 8- and 6-bit RDN$\times$2, DFSQ obtains 32.915 dB and 32.184 dB PSNR on BSD100, corresponding to 0.002 dB and 0.013 dB drop. While on 8- and 6-bit RDN$\times$4, DFSQ only incurs decreases of 0.002 dB and 0.011 dB on BSD100, respectively. 
Also, the performance advantage of our DFSQ becomes increasingly apparent as the bit-widths decrease.
In particular, for RDN$\times$2 on Urban100, DFSQ improves the PSNR by 0.027 dB, 0.104 dB, 0.176 dB, and 0.418 dB on 8-, 6-, 4-, and 3-bit, respectively. 
While for RDN$\times$4 on Urban100, our DFSQ obtains performance gains by 0.03 dB, 0.109 dB, 0.171 dB, and 0.235 dB PSNR on 8-, 6-, 4-, and 3-bit, respectively.
Moreover, it can be observed that the optimization-based methods do not even give higher results than min-max methods for 6- and 4-bit cases, indicating the existence of the over-fitting issue.

\section{Discussion}

Despite our DFSQ makes big progress, it involves an expensive channel-wise normalization before quantization. Thus, reducing the overhead incurred by normalization is worth to be further explored. For example, the de-normalization can be realized in low-bit as in \cite{dai2021vs,hong2022daq}.  
In addition, although optimization-based methods exhibit satisfactory performance on high-level tasks, they suffer from the over-fit issue as shown in Sec.\,\ref{sec:quantitative}. Therefore, a specialized optimization-based PTQ for SR could be a valuable direction.

\section{Conclusion}
In this paper, we present a novel quantization method, termed Distribution-Flexible Subset Quantization (DFSQ) for post-training quantization on super-resolution networks. We discover that the activation distribution of SR models exhibits significant variance between samples and channels. Correspondingly, our DFSQ suggests conducting a channel-wise normalization for activation at first, then applying the hardware-friendly and distribution-flexible subset quantization, in which the quantization points are selected from a universal set consisting of multi-word additive log-scale values.
To select quantization points efficiently, we propose a fast quantization points selection strategy with linear time complexity. We perform $K$-means clustering to identify the closest quantization points to centroids from the universal set.
Our DFSQ shows its superiority over many competitors on different quantized SR models across various bit-widths and benchmarks, especially when performing ultra-low precision quantization.

\noindent\textbf{Acknowledgements}. This work was supported by National Key R\&D Program of China (No.2022ZD0118202), the National Science Fund for Distinguished Young Scholars (No.62025603), the National Natural Science Foundation of China (No. U21B2037, No. U22B2051, No. 62176222, No. 62176223, No. 62176226, No. 62072386, No. 62072387, No. 62072389, No. 62002305 and No. 62272401), and the Natural Science Foundation of Fujian Province of China (No.2021J01002,  No.2022J06001).

\renewcommand\refname{References}
\bibliographystyle{plain}
\bibliography{main}

\clearpage
\appendix
\section*{Appendix \label{appendix}}

\begin{figure*}[!htbp]
\centering
\subfloat[
{\centering \label{fig:hr-result-edsrx4-1}}]
{
\centering\includegraphics[width=0.85\linewidth]{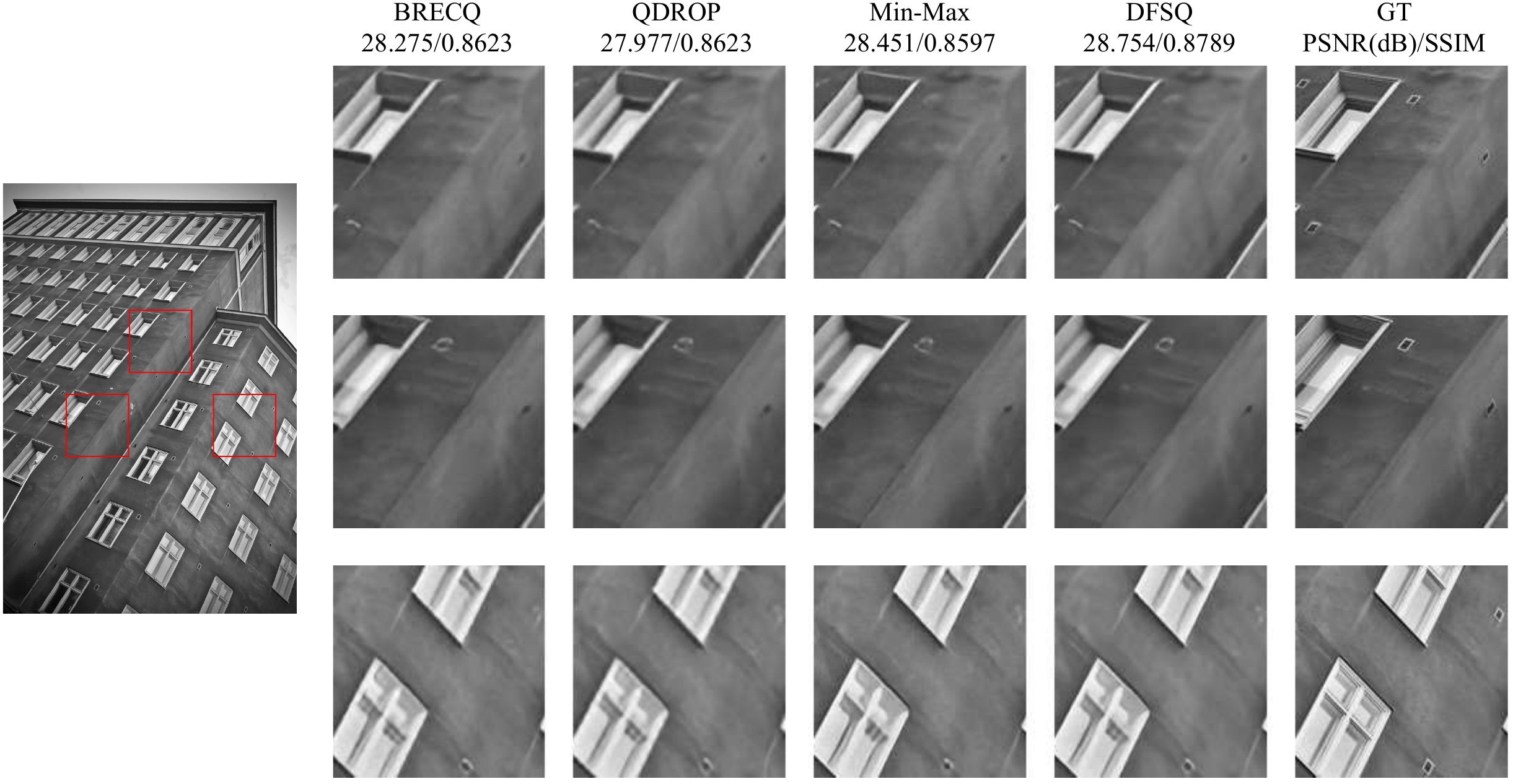} \hfill
}
\\
\subfloat[
{\centering \label{fig:hr-result-edsrx4-2}}]
{
\centering\includegraphics[width=0.85\linewidth]{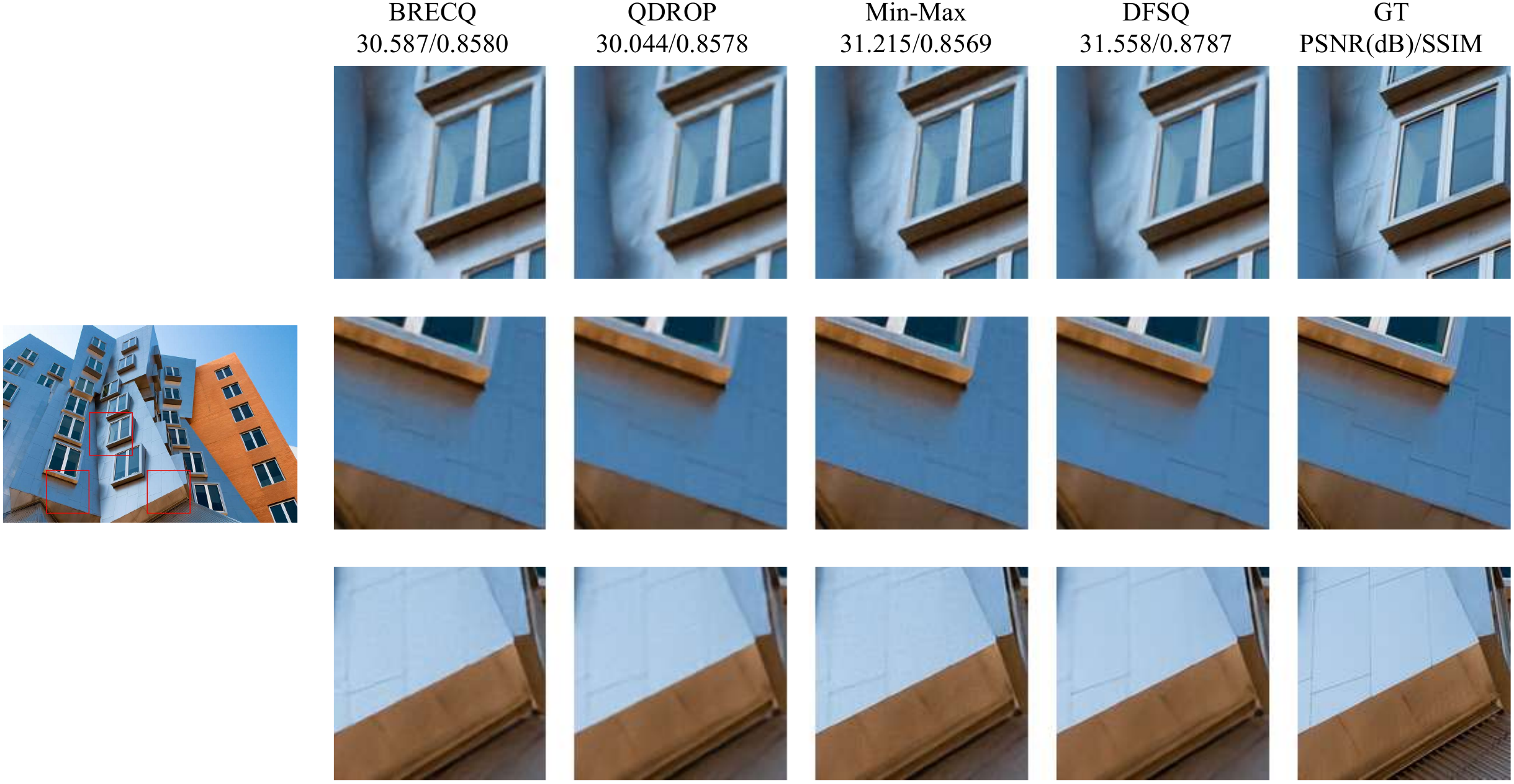} \hfill
}
\caption{Illustration of the qualitative results of 4-bit EDSR$\times$4.}
\label{fig:esdr}
\end{figure*}

\begin{figure*}[!htbp]
\centering
\subfloat[
{\centering \label{fig:hr-result-rdnx4-1}}]
{
\centering\includegraphics[width=0.85\linewidth]{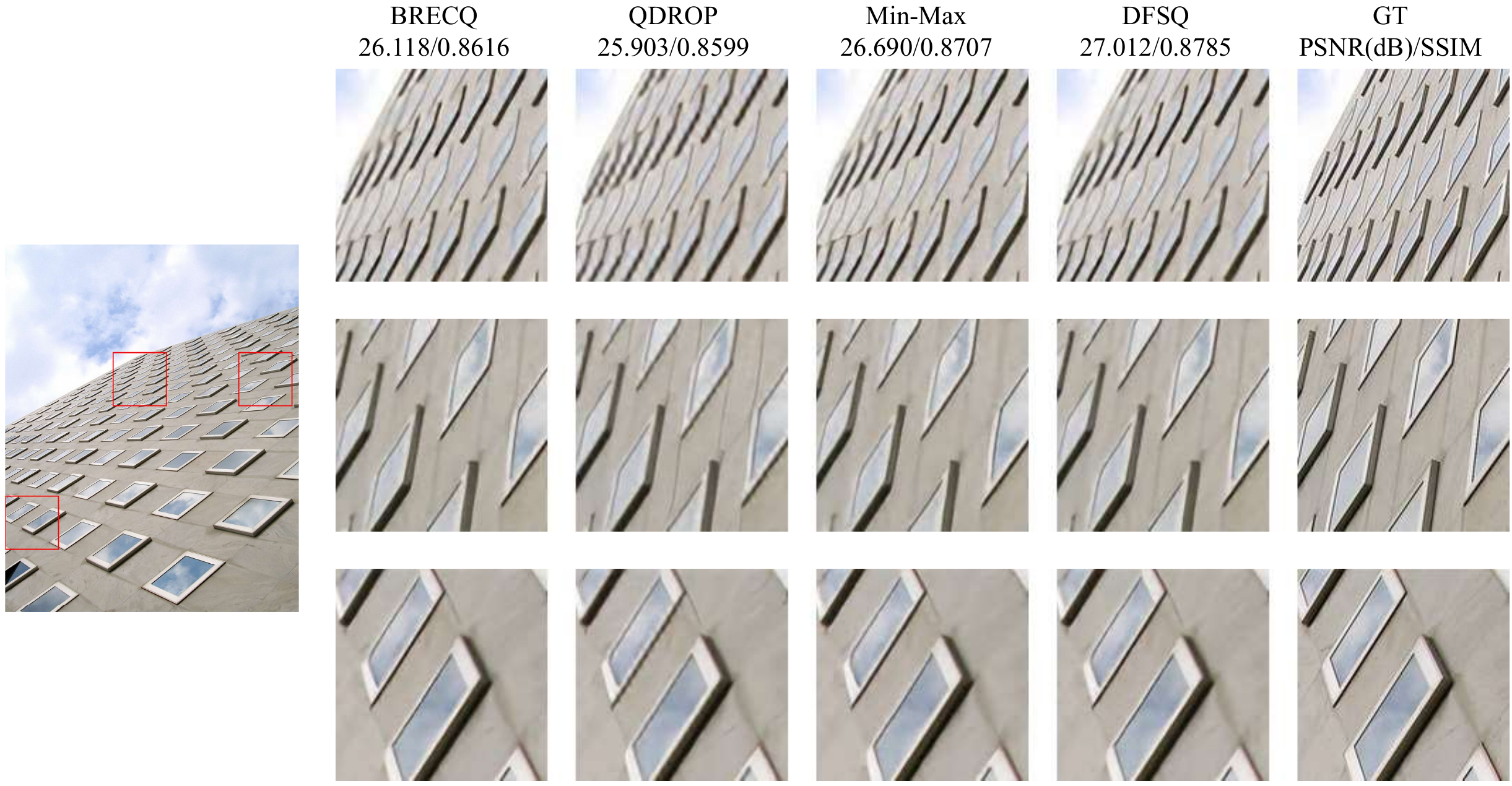} \hfill
}
\\
\subfloat[
{\centering \label{fig:hr-result-rdnx4-2}}]
{
\centering\includegraphics[width=0.85\linewidth]{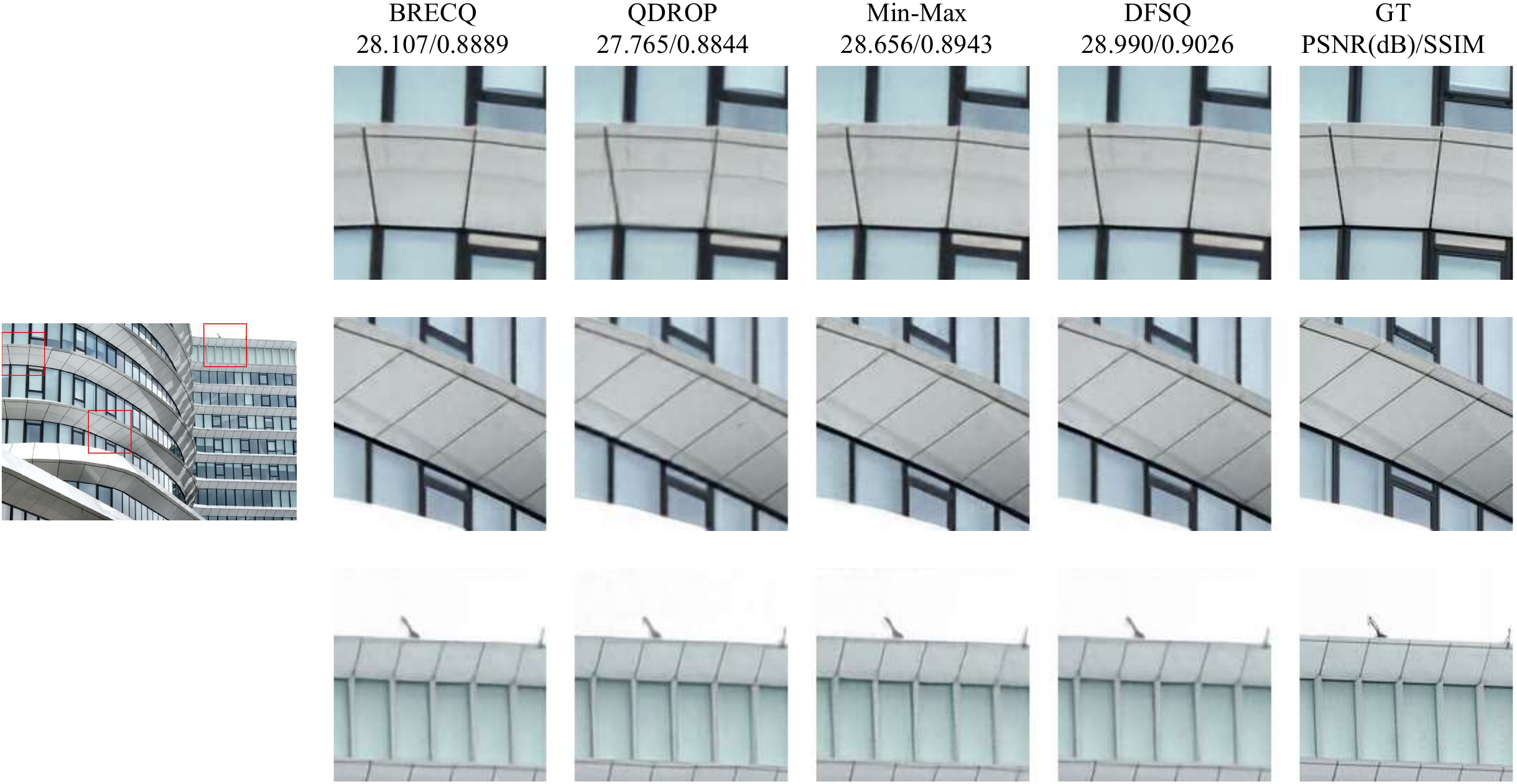} \hfill
}
\caption{Illustration of the qualitative results of 4-bit RDN$\times$4.}
\label{fig:rdn}
\end{figure*}

\begin{figure*}[!htbp]
\centering
\subfloat[
{\centering \label{fig:insight-RDNx4_channel}}]
{
\centering\includegraphics[width=0.85\linewidth]{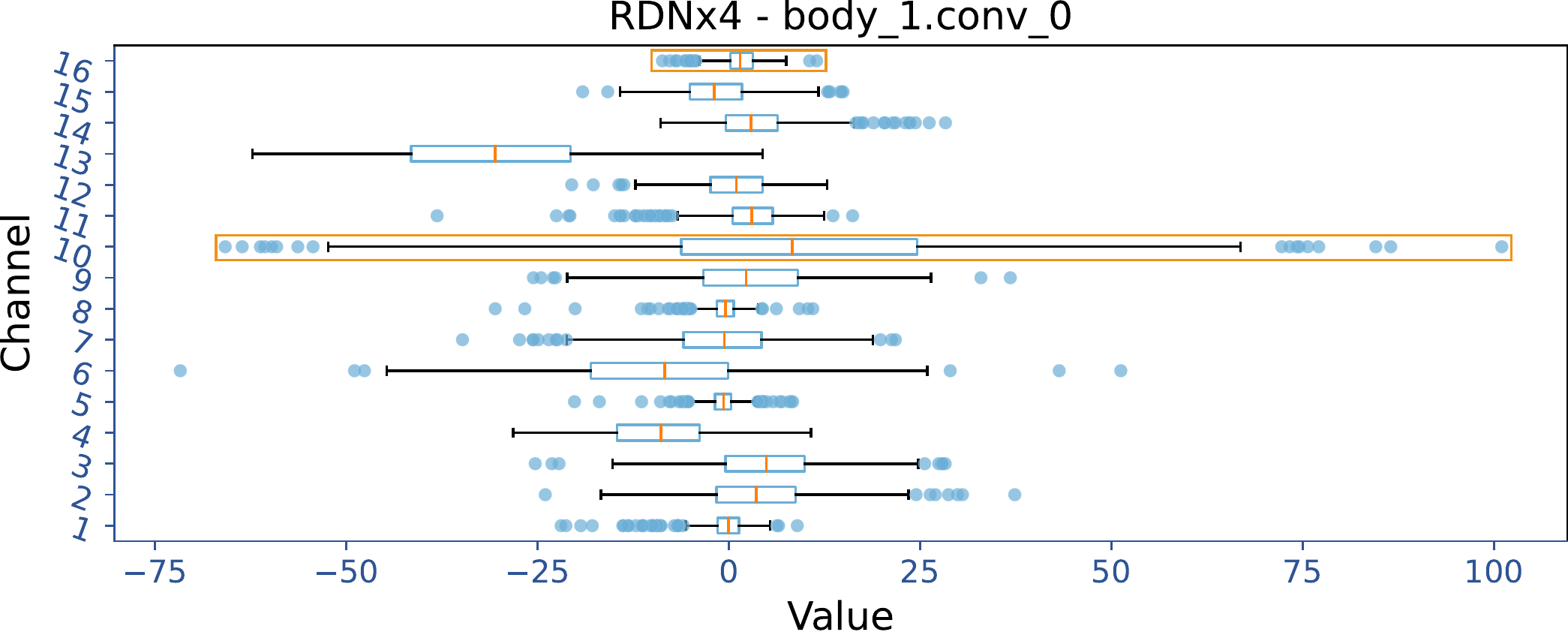} \hfill
}
\\
\subfloat[
{\centering \label{fig:insight-RDNx4_sample}}]
{
\centering\includegraphics[width=0.85\linewidth]{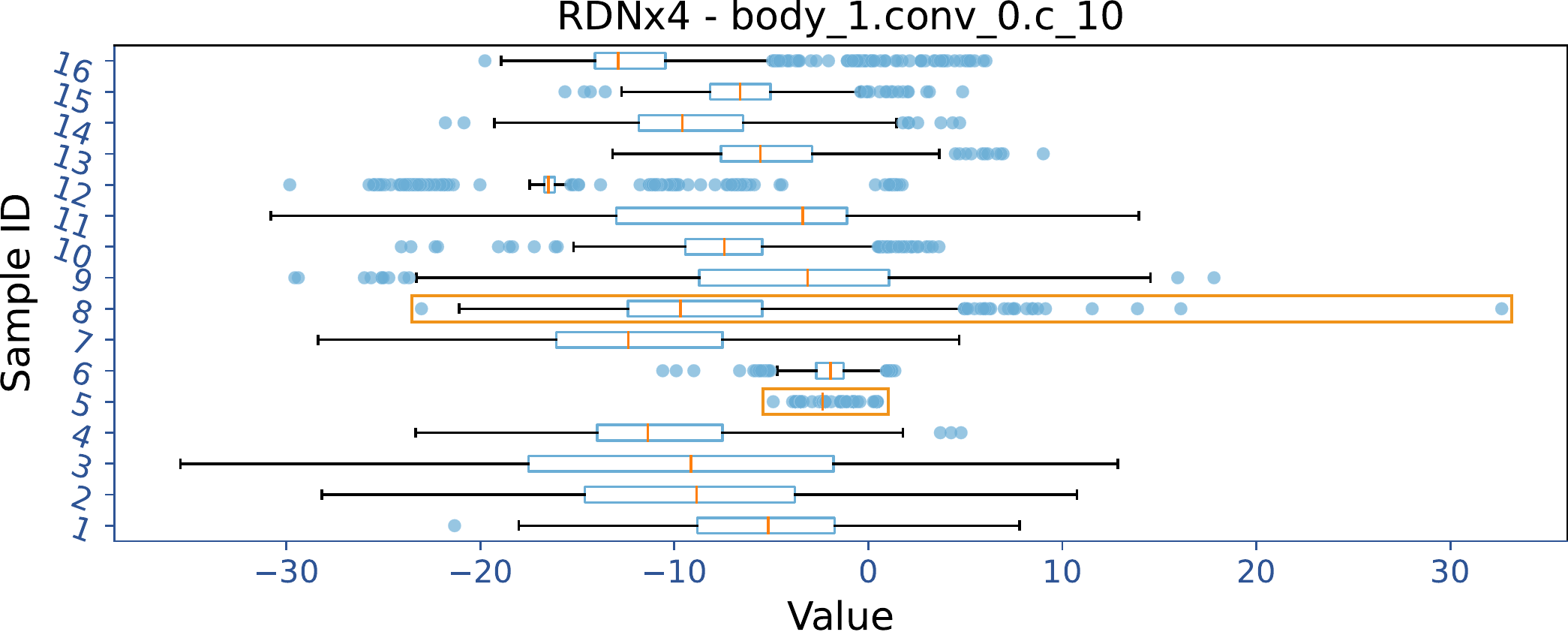} \hfill
}
\caption{Illustration of high variance activation within RDN$\times$4. (a) and (b) present the activation distribution of different channels given the same sample and of different samples given the same channel, respectively. The orange box depicts the data with the maximum discrepancy.}
\label{fig:insight-rdn}
\end{figure*}

\section{Qualitative Results}

Fig.\,\ref{fig:esdr} and Fig.\,\ref{fig:rdn} exhibit the qualitative results of the 4-bit EDSR$\times$4 and 4-bit RDN$\times$4, respectively. The reported PSNR/SSIM are measured by the displayed image. It can be seen that our method obtains the best visualization results compared with other methods, which demonstrates the superiority of our DFSQ.

\section{More Illustrations}

In this section, we provide more illustrations of the activation within SR models. Fig.\,\ref{fig:insight-rdn} provides the activation distribution of RDN$\times$4. It can be seen that the distribution of activation varies a lot across different channels and samples. For example, as shown in Fig.\,\ref{fig:insight-RDNx4_channel}, the range of channel\_16 is -10 to 10, while the range of channel\_16 is -65 to 100. Distributions of the same sample but different channels shown in Fig.\,\ref{fig:insight-RDNx4_sample} also exhibit a large variance. 
Specifically, the range of channel\_5 is -6 to 2, while the range of channel\_8 is -23 to 32.

\begin{table*}[ht]
\centering
\caption{Universal set designing.}
\begin{tabular}{c|cccc}
\toprule[1.25pt]
Settings     & Setting1                                           & Setting2                                            & Setting3                                            & Setting4                                            \\ \hline\hline
Word Sets & \begin{tabular}[c]{@{}c@{}}\{1, $2^{-1}, 2^{-3}, 0$\},\\ \{1, $2^{-2}, 2^{-4}, 0$\}. \end{tabular} & \begin{tabular}[c]{@{}c@{}} \{1, $2^{-1}, 2^{-3}, 0$\},\\ \{1, $2^{-2}, 2^{-4}, 0$\},\\ \{1, $2^{-3}, 2^{-5}, 0$\}. \end{tabular} & \begin{tabular}[c]{@{}c@{}} \{1, $2^{-1}, 2^{-5}, 0$\},\\ \{1, $2^{-2}, 2^{-6}, 0$\}, \\ \{1, $2^{-3}, 2^{-7}, 0$\}, \\ \{1, $2^{-4}, 2^{-8}, 0$\}. \end{tabular} & \begin{tabular}[c]{@{}l@{}} \{1, $2^{-1}, 2^{-6}, 0$\},\\ \{1, $2^{-2}, 2^{-7}, 0$\}, \\ \{1, $2^{-3}, 2^{-8}, 0$\}, \\ \{1, $2^{-4}, 2^{-9}, 0$\}, \\ \{1, $2^{-5}, 2^{-10}, 0$\}. \end{tabular} \\ \bottomrule[0.75pt]
\end{tabular}
\end{table*}

\begin{figure*}[htbp]
\centering
\includegraphics[width=0.7\linewidth]{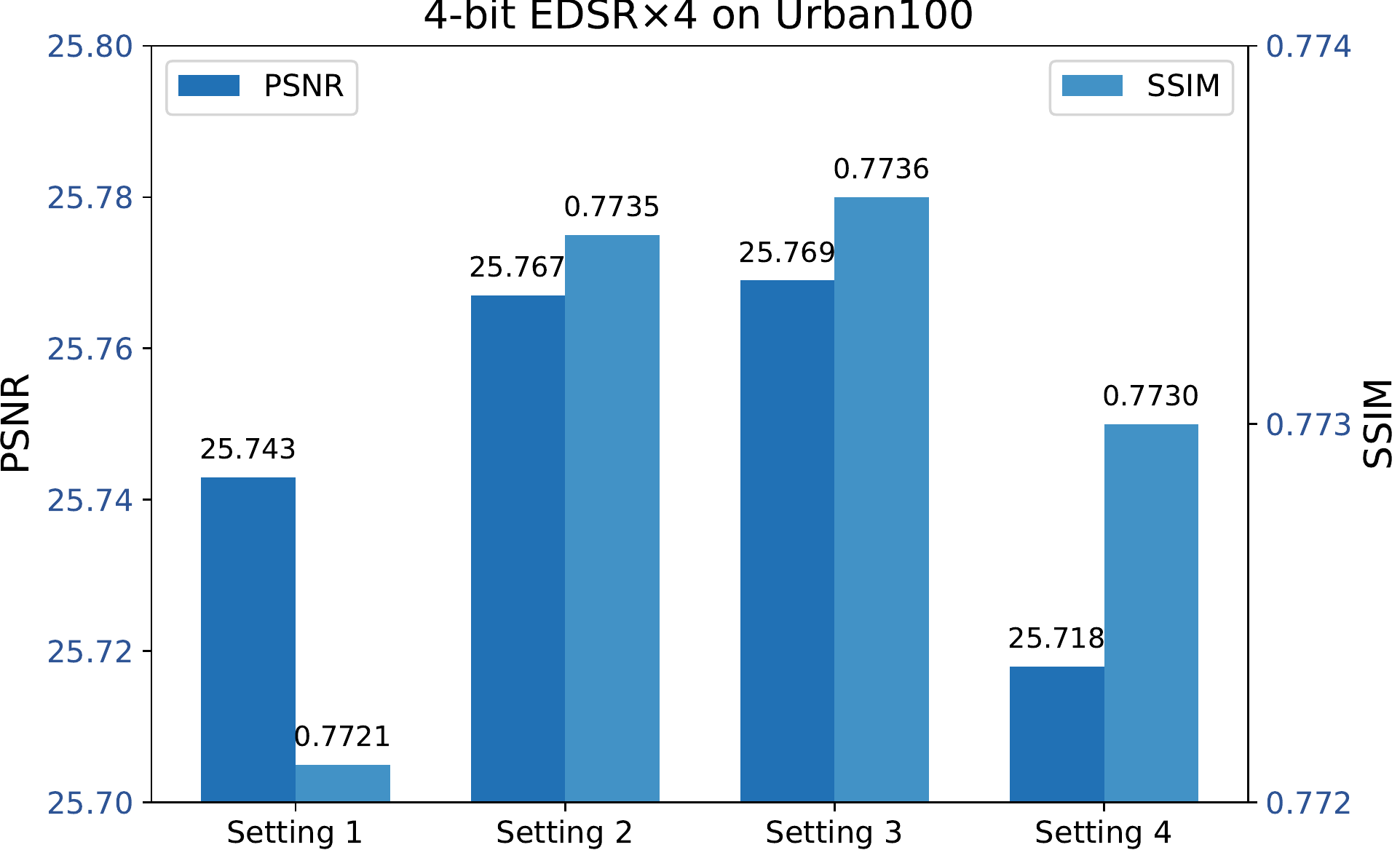} 
\caption{Results of 4-bit EDSR$\times$4 on Urban100 with different universal set settings.}
\label{fig:abl_univ_set}
\end{figure*}

\section{Ablation of Universal Set}

In this section, we provide the experimental results of different universal sets by adjusting the number of word sets. In particular, we fix the size of each word set and vary the number of word sets to construct different universal sets.
Four settings including 2$\times$4, 3$\times$4, 4$\times$4, and 5$\times$4 are provided as presented in Tab.\,\ref{fig:insight-rdn}. Note that the 4$\times$4 setting is the one we used in our main paper.
The performance comparison is presented in Fig.\,\ref{fig:abl_univ_set}. It can be seen that by increasing the number of word sets from 2 to 3 (2$\times$4 \emph{vs.} 3$\times$4), the PSNR is improved by 0.34 dB. When the number of word sets is further increased to 4, the PSNR is improved by 0.002 dB. While at the 5$\times$4 setting, the PSNR drops by 0.051 dB, indicating the over-fitting issue.  
Thus, we choose to use the 4$\times$4 setting since the division of 4 can be achieved by performing bit shift operations on the elements of each word set.

\end{document}